# A One-class Classification Framework using SVDD : Application to an Imbalanced Geological Dataset

Soumi Chaki[1], Akhilesh Kumar Verma[2], Aurobinda Routray[1], William K. Mohanty[2], Mamata Jenamani[3]

[1]Department of Electrical Engineering,
IIT Kharagpur
Kharagpur, India
soumibesu2008@gmail.com
aroutray@ee.iitkgp.ernet.in

[2] Department of Geology and Geophysics,
IIT Kharagpur
Kharagpur, India
akhileshdelhi2007@gmail.com
wkmohanty@gg.iitkgp.ernet.in

[3]Department of Industrial and Systems Engineering,
IIT Kharagpur
Kharagpur, India
mj@iem.iitkgp.ernet.in

*Abstract*— Evaluation of hydrocarbon reservoir requires classification of petrophysical properties from available dataset. However, characterization of reservoir attributes is difficult due to the nonlinear and heterogeneous nature of the subsurface physical properties. In this context, present study proposes a generalized one class classification framework based on Support Vector Data Description (SVDD) to classify a reservoir characteristic– water saturation into two classes (Class high and Class low) from four logs namely gamma ray, neutron porosity, bulk density, and P-sonic using an imbalanced dataset. A comparison is carried out among proposed framework and different supervised classification algorithms in terms of g-metric means and execution time. Experimental results show that proposed framework has outperformed other classifiers in terms of these performance evaluators. It is envisaged that the classification analysis performed in this study will be useful in further reservoir modeling.

## I. INTRODUCTION

In the process of reservoir quantification for the production of hydrocarbon, there are several challenges to be solved. These issues include classification of different lithological units, integration of different types of data recorded in different domain, problem of non-uniform sampling, heterogeneous characteristics of reservoir variables, etc. Heterogeneity, i.e. non-uniform, nonlinear characteristics of reservoir properties, introduces difficulty in reservoir modeling. These modeling are carried out using state-of-art nonlinear approaches such as Artificial Neural Networks (ANN), Fuzzy Logic (FL), Genetic Algorithm (GA), etc. Some applications of these methods in the field of petroleum reservoir modeling are discussed in [1]–[4]. However, it has been observed that the accuracy in reservoir modeling can be improved using classification based approaches [5]. Thus, classification of petrophysical parameters is beneficial for reservoir studies. Now, it is a complex task whose performance depends on the available subsurface information. Supervised classifiers are generally selected over unsupervised clustering algorithms due to the complex nature of the problem. Nevertheless, the requirement of a complete and representative training dataset is must for accurate learning of these supervised classifiers. In case of an imbalanced dataset, these constraints of the training dataset do not get satisfied. Moreover, the underrepresented training dataset may have several class distribution skews. Recently, the learning problems from imbalance dataset have received interest from researchers due to existence of such dataset in "real-world applications" [6]–[9]. Kernel based methods have gained acceptance in classification of imbalanced dataset over other supervised classification methods, especially in remote sensing fields [10]–[12]. Support vector data description (SVDD) is a latest kernel based algorithm which has attracted attention from researchers of different fields for its ability in learning without any a priori knowledge on distribution of dataset [13]–[15].

The first important contribution of this paper is to propose a generalized framework based on Support Vector Data Description (SVDD) [13], [14] to characterize water saturation from input well logs. Next, a comparative analysis is presented to demonstrate the effectiveness of the proposed classification method over other classifiers (discriminant [16], [17], naive Bayes [16], [18], support vector machine based classifier [19], [20]). A dataset from four closely spaced wells are selected for this study. Here, combined dataset of three wells are used for training, and remaining one well is used for testing.

The rest of the paper is structured as follows: first, the data used in this study is described; next, the theory of SVDD is briefly presented; after that the proposed classification framework is described. Then, a brief description of performance evaluators used in this work is given. In the following section, experimental results are reported. Finally, we conclude this paper with the discussion and future scope.

## II. DESCRIPTION OF DATASET

The well logs used in this work are acquired from four closely spaced boreholes located in an onshore hydrocarbon field of India. Henceforward, these aforementioned wells are to be referred as A, B, C, and D, respectively. The borehole data contains several logs such as gamma ray content (GR), bulk density (RHOB), P-sonic (DT), neutron porosity (NPHI), spontaneous potential (SP) and different resistivity logs such as deep resistivity (RT), medium resistivity (RM) and shallow resistivity (RS) logs. Reservoir characteristics, e.g., sand fraction, porosity, water saturation, oil saturation etc. are derived from these log properties. Literature study reveals that gamma ray content (GR), bulk density (RHOB), P-sonic (DT), neutron porosity (NPHI), spontaneous potential (SP)are among different logs to be used as predictor variables to model or classify lithological properties. After selection of relevant features among available logs, we have used gamma ray content (GR), bulk density (RHOB), P-sonic (DT), and neutron porosity (NPHI) logs as input attributes to classify water saturation level. The rock properties of subsurface formations can be interpreted from these variables. The gamma radiation of different formations along the depth is represented by gamma ray log in American Petroleum Institute (API) unit. The density log is recorded in grams per cubic centimeter unit. It varies according to mineralogy and porosity values. Travel time of P-waves versus depth is recorded as P-sonic log in micro second per feet. The fourth predictor variable i.e. neutron porosity log is attuned to read the true porosity and represented in per unit. In this work, the target variable is water saturation, which is an important characteristic in the petroleum industry representing the fraction of formation water present in the pore space.

III. SUPPORT VECTOR DATA DESCRIPTION

Large dataset can be characterized using data description techniques. Significant efforts have been made for the classification of real world datasets. Support Vector Data Description (SVDD), an extension of Support Vector Machine (SVM), is widely used approach for the data classifications [13], [14].

In general, data are described by defining a close boundary around the data. This close boundary is defined by hypersphere, $F(R,a)$ where '$a$' represents center and '$R$' is the radius. Volume of the hypersphere should be minimized for the data description [13]–[16]. Outlier in the data can be characterized by defining slacks variables $\varepsilon_i \geq 0$. In this case, the minimization term of error function is given by

$$F(R,a) = R^2 + C\sum_i \varepsilon_i \|x_i - a\|^2 \leq R^2 + \varepsilon_i \quad (1)$$

where,

$$\|x_i - a\|^2 \leq R^2 + \varepsilon_i, \text{ for all } i \quad (2)$$

Kernel function $K(x_i, x_j) = \phi(x_i).\phi(x_j)$ is used for smooth data description. Then the SVDD function can be represented as

$$L = \sum_i \alpha_i K(x_i, x_j) - \sum_{i,j} \alpha_i \alpha_j K(x_i, x_j)$$
$$\text{for all } \alpha_i : 0 \leq \alpha_i \leq C \quad (3)$$

Optimization of equation 3 gives the data description which can be obtained by several algorithms available in the literature, and Lagrange multipliers should satisfy the normalization constraint $\sum_i \alpha_i = 1$. The values of $\alpha_i s$ can be found out by minimizing $L$. We have used a Gaussian kernel

$$K(x_i, x_j) = e^{-q\|x_i - x_j\|} \quad (4)$$

to represent the dot product $\phi(x_i).\phi(x_j)$ as discussed in [15], [16], [21]. In order to calculate the radius we have to look for the support vectors. Firstly, $R^2(x)$ in terms of the kernel function for each of the point is found out. Then, we get

$$R^2(x) = K(x,x) - 2\sum_i \alpha_i K(x_i, x)$$
$$+ \sum_{i,j} \alpha_i \alpha_j K(x_i, x_j) \quad (5)$$

Now the support vectors are those data objects which lie on the surface of the hypersphere i.e., for which $C = \alpha_i$. The contours formed due to the data points are cluster boundaries. For the purpose of our work, we take the radius of the circle $R$ to be the maximum of values $R(x)$ for the support vectors. Any data point lying beyond $R$ is considered to be an outlier. In one class classification using SVDD, the minority class patterns are used as target in training phase to construct the hypersphere. Once the hypersphere is constructed, the classifier is evaluated using majority class patterns as testing dataset. For imbalanced dataset, the improvement in one-class classifier performance compared to its two-class counterpart is apparent [22]–[23].

IV. PROPOSED CLASSIFICATION FRAMEWORK

In the recent years, SVDD and other kernel based algorithms have been reported as popular techniques adapted for classification of imbalanced dataset in the field of hyperspectral image processing, outlier detection, document classification etc. In this work, an attempt has been made to construct a SVDD based framework to classify reservoir properties using an imbalanced geological dataset. The proposed generalized framework, which includes three steps namely- 1) data preparation, 2) preliminary analysis, and 3) training and testing, is represented in Fig. 1. These steps are briefly discussed in this section.

*A. Data Preparation*

Well log data from four wells located in the western onshore hydrocarbon field of India are used in the present study. The procedure of data preparation is started with data acquisition as shown in Fig 1. The log files contain a number of missing data values. These patterns are removed to make a data file of valid values only. Then we uniformly re-sampled the data.

*B. Preliminary Analysis*

Feature selection plays a crucial role in tuning the performance of pattern classifiers. In the pre-processing stage, several number of "candidate features" are extracted from raw dataset. Then relevant features are selected using different algorithms i.e. mutual information, Relief algorithm, and its variants. Here, we use Relief algorithm [24], which identifies statistically relevant features and performs well in case of noisy dataset, to select input attributes before starting to train the classifier. Designing a classifier with several inputs prolongs training time along with unnecessary proliferation in the model complexity. Moreover, the generalization capability of a model enhances while using only relevant features as inputs.

Next, we classify the water saturation into two classes, namely- Class high and Class low using a user defined threshold. The choice of threshold value is guided by two factors. Firstly, saturation values belonging to the Class high must be as close to one as possible while in Class low it must be as close to zero as possible. This is done by observing the histogram of the saturation values. Secondly, the high computational complexity of the SVDD classifier has compelled us to set the threshold in a manner so as to have reasonable small number of patterns at least in one class to have the classifier trained within reasonable time. This threshold value is modified depending on the training speed of the SVDD algorithm. After completion of the preliminary analysis, training and testing of SVDD based one class classifier is started. Besides, selection of the threshold level is confirmed by expert geologists.

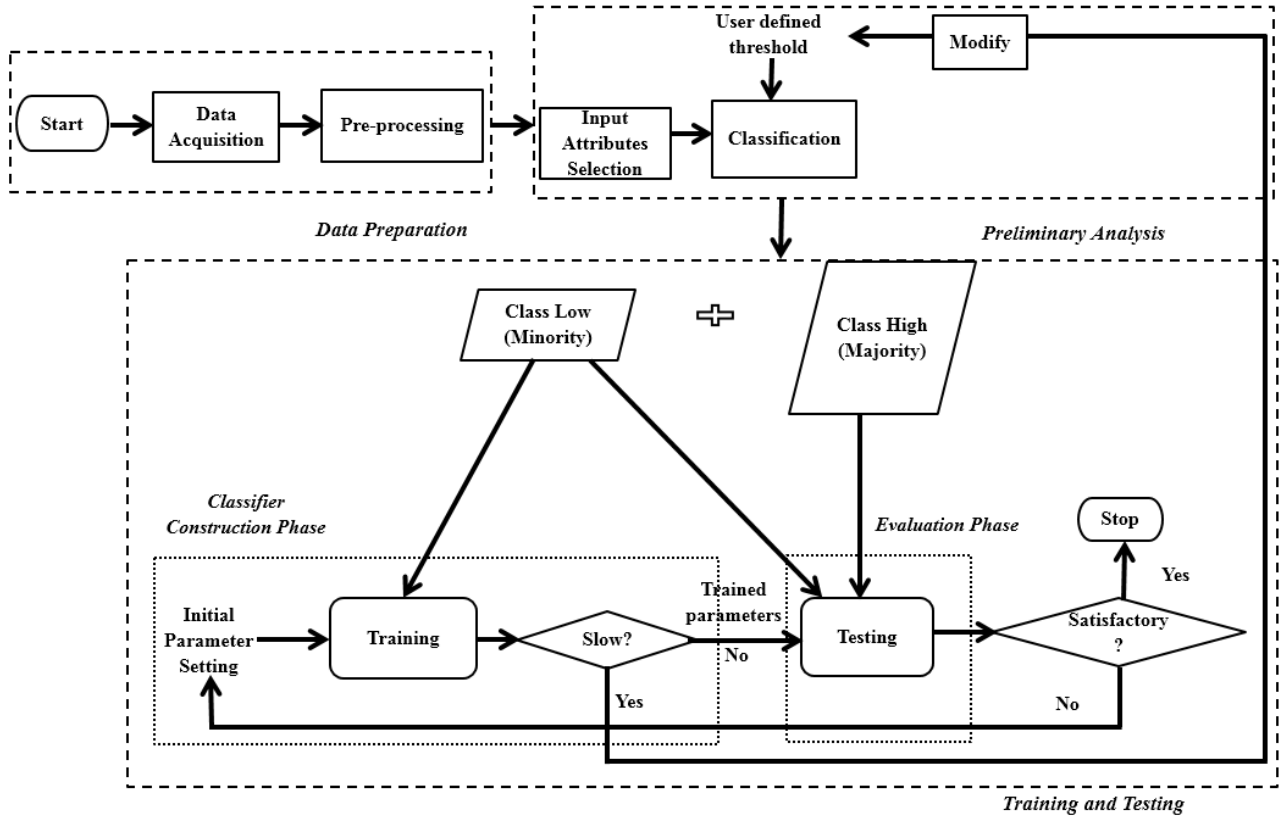

Fig. 1. Proposed framework for classification of imbalanced dataset

*C. Training and Testing*

The training and testing steps associated with the one-class classifier are shown in the bottom part of Fig. 1. In this problem, the available patterns are significantly large in case of Class high compared to Class low. In other words, Class high and Class low can be invariably denoted as majority and minority classes. In this case study, one well is used for blind testing. The patterns belong to minority classes of integrated dataset of three well logs are used in the training phase. The minority class patterns of the remaining fourth well along with majority class patterns of the four wells are used to test the classifier performance.

The input attributes (gamma ray, neutron porosity, bulk density, and P-sonic log) of training patterns are used to construct the SVDD hypersphere. Classification accuracy of SVDD is improved by adjusting few parameters: type of the kernel function and associated parameters, and radius of the hypersphere $C$. The kernel functions such as Gaussian, higher order polynomial (2–10), radial basis function, exponential radial basis function, kernel parameters, are experimented with values of $C$ varying from 0 to 1. The classifier uses a Lagrangian function which is minimized using constrained optimization. It divides the patterns into two classes as true data which resides inside the hypersphere and outliers which reside outside the boundary of the hypersphere. The points which make the boundary of the hypersphere are called support vectors. In this work, we include these support vectors in the outlier class. The trained parameters are saved and applied to majority class to test the classifier performance.

After completion of the training and testing stage, the classification performance achieved using this proposed framework is compared to other classifiers namely discriminant, naive Bayes, and support vector machine based classifier.

## V. PERFORMANCE EVALUATORS

The performance of the proposed framework using one class classifier based on SVDD is evaluated upon the accuracy of both positive and negative classes. Instead of employing confusion matrix, which is generally used to measure performance of classifier, here we use g-metric means [25]. This performance evaluator is often used in case of imbalanced dataset. G-metric means can be represented as

$$g=\sqrt{acc_P * acc_N} \qquad (6)$$

where $acc_P$ and $acc_N$ represent sensitivity and specificity, respectively. Sensitivity indicates the accuracy on the positive instances i.e. (true positives/ (true positives + false negatives)) and similarly, specificity denotes the accuracy on the negative instances i.e. (true negatives/ (true negatives + false positives)).

Program execution time is also recorded to compare the performance of proposed framework with respect to other classifiers.

## VI. EXPERIMENTAL RESULTS

The experiments carried out in this work are performed on MATLAB platform on a Intel(R) Core(TM) i5-2410M CPU @2.30 GHz workstation having 4 GB RAM. The experimental results and analysis are reported in this work according to the respecting sections.

### A. Dataset Preparation

This stage is the starting point of the proposed framework. Well logs are selected and pre-processed.

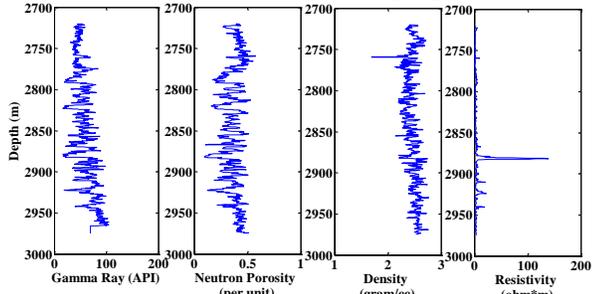

Fig. 2: Plots of gamma ray, neutron porosity, bulk density, and resistivity along depth for well A

Fig. 2 represents plots of gamma ray, neutron porosity, bulk density, and resistivity logs along depth for well A. Similarly, Fig. 3 represents P-sonic, acoustic impedance, and water saturation logs along depth for the same well. Designing a classifier is required to classify water saturation log from available log variables. The selection of the input variables is carried out using Relief algorithm as discussed in the following section.

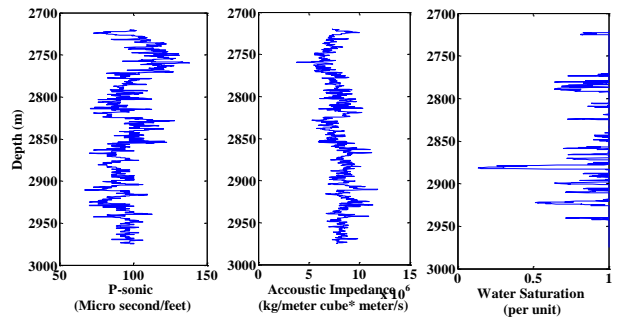

Fig. 3: Plots of P-sonic, acoustic impedance, and water saturation logs along depth for well A

### B. Preliminary Analysis

First, several attributes are extracted from raw dataset. Then, four relevant attributes are selected from the six "candidate attributes" using Relief algorithm. The result of Relief algorithm is represented in Fig. 4. It can be observed from the figure that gamma ray (GR), neutron porosity (NPHI), bulk density (RHOB), and P-sonic (DT) logs are more relevant features related to water saturation in terms of predictor importance weight compared to deep resistivity (RT) and acoustic impedance (DT) logs.

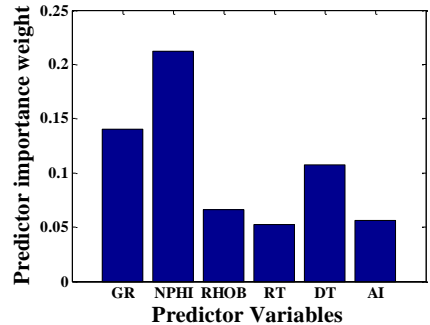

Fig. 4: Selection of relevant input attributes using Relief algorithm

After selection of appropriate input attributes the next task is to classify water saturation into two classes using user defined threshold value. We consider two criterion as discussed in the earlier section for the selection of the threshold level to classify the water saturation values into two classes. For this particular problem, we choose 0.7 as the threshold value after verifying the constraints related to computational speed of SVDD algorithm and experience geoscientists' view. Patterns with saturation level greater than or equal to 0.7 are called Class high and the other patterns are called Class low. We have 3% of the whole data set in the Class low set. It can be observed from Fig. 5 that the distribution of water saturation values is skewed at one. Specifically, 97% of the total available patterns belong to Class high which is associated with higher values of water saturation. Therefore, Class low and Class high can be termed as minority and majority classes respectively.

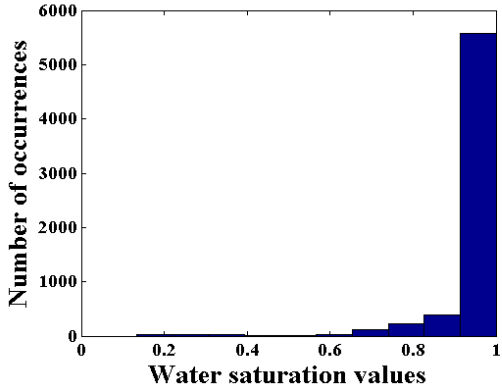

Fig. 5 : Histogram plot for water saturation

## C. Training and Testing

As discussed in the methodology section, the classifier is constructed and trained using the input attributes of minority class patterns with an initial parameter setting. After optimizing the Lagrangian function using constrained optimization, the trained parameters are used to examine the classifier capability using testing dataset. First, the data of one well is set aside. After training the SVDD using combined minority class patterns of remaining three wells, we test the performance of the classifier using majority class patterns of these three wells along with all patterns (majority and minority) of test well. The results reported in this article corresponds to blind testing of individual well when classifier learning is carried out using a kernel function and initial $C$ value. For example, in case of blind prediction of well C, the SVDD hypersphere is constructed using patterns belong to minority class from combined dataset of remaining three wells using Gaussian kernel of width parameter of 2.0, and $C = 0.008$ as initial parameter setting.

SVM, naïve base, and discriminant classifiers are optimized after initializing with appropriate parameter values using the same predictor variables. From the test output, the patterns classified as outliers and support vectors are considered to be majority class components; and data vectors are specified as minority class components. Then, comparison is carried out among these supervised classifiers depending on blind testing result of each of the wells. Table I and II represent comparison result of proposed framework with other supervised classifiers.

TABLE I: PERFORMANCE COMPARISON OF CLASSIFIERS IN TERMS OF G-METRIC MEAN

| Well Name | Value of g-metric mean | | | |
|---|---|---|---|---|
| | SVM | Naive Bayes | Discriminant | Proposed Workflow (SVDD) |
| A | 0.75 | 0.75 | 0.70 | 0.78 |
| B | 0.61 | 0.50 | 0.59 | 0.65 |
| C | 0.71 | 0.81 | 0.80 | 0.83 |
| D | 0.74 | 0.80 | 0.68 | 0.90 |
| Average Performance | 0.70 | 0.71 | 0.69 | 0.79 |

TABLE II: PERFORMANCE COMPARISON OF CLASSIFIERS IN TERMS OF PROGRAM EXECUTION TIME (SECONDS)

| Well Name | Program execution time (in seconds) | | | |
|---|---|---|---|---|
| | SVM | Naive Bayes | Discriminant | Proposed Workflow (SVDD) |
| A | 50.0 | 44.1 | 32.3 | 30.2 |
| B | 40.2 | 34.0 | 43.1 | 40.5 |
| C | 43.1 | 30.1 | 45.1 | 19.3 |
| D | 43.1 | 54.7 | 40.3 | 26.4 |
| Average Performance | 44.1 | 40.7 | 40.2 | 29.1 |

It is evident from the tables I and II that the proposed classifier workflow outperformed other supervised classifiers in terms of g-metric means and program execution time.

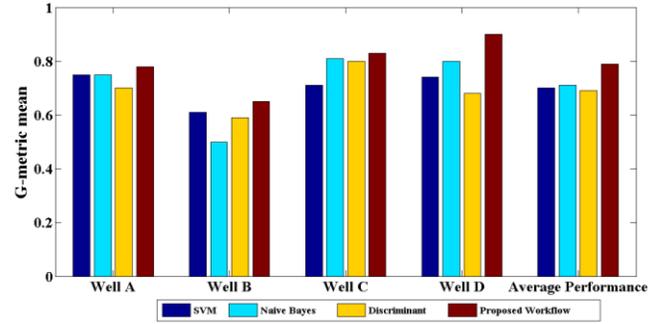

Fig. 6 : Bar plot describing performance of classifiers in terms of g-metric means

Fig. 6 and 7 represent the result of performance comparison of supervised classifiers in terms of g-metric means and program execution time respectively.

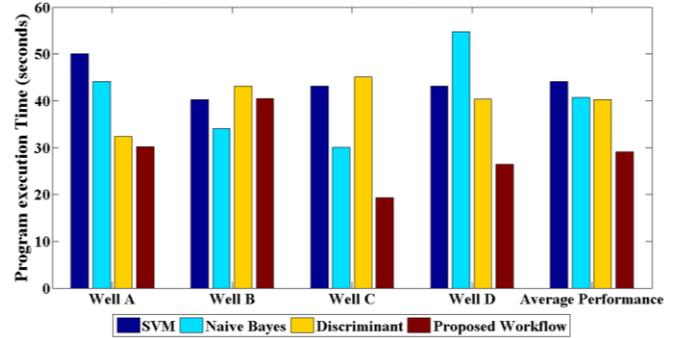

Fig. 7 : Bar plot describing performance of classifiers in terms of program execution time

Therefore, it can be inferred from the results that the proposed workflow based on SVDD can be used as a powerful tool to classify imbalanced dataset in reservoir characterization domain.

## VII. CONCLUSION AND FUTURE SCOPE

In this work, a complete framework based on SVDD is proposed to classify water saturation from well logs using an imbalanced geological dataset. Comparative analysis reported in this paper has shown that the proposed methodology outperformed existing classifier algorithms in terms of performance evaluators. This work can be extended with inclusion of seismic attributes as inputs to the classifier based model. Integration of seismic and limited number of available borehole data will help to produce 3D volume representing high and low water saturation values throughout a study area. Efforts can be made to improve the speed of the algorithm.